\documentclass{article}
\usepackage{ijcai20}

\usepackage{times}

\usepackage{soul}
\usepackage{url}
\usepackage[hidelinks]{hyperref}
\usepackage[utf8]{inputenc}
\usepackage[small]{caption}
\usepackage{graphicx}
 \usepackage{amssymb}
\usepackage{amsmath}
 \usepackage{amsthm}
  \usepackage{mathrsfs}
\usepackage{booktabs}
\title{Learning to Generate Time Series Conditioned Graphs with Generative Adversarial Nets}

\author{
Shanchao~Yang$^1$\footnote{Contact Author}\and
Jing~Liu$^2$\and
Kai~Wu$^3$\And
Mingming~Li$^4$\\
\affiliations
$^{1,2,3,4}$Xidian University\\
\emails
yangshanchaoysc@gmail.com,
neouma@163.com,
kaiwu@stu.xidian.edu.cn,
at.mingli@gmail.com
}

\begin{document}

\maketitle

\begin{abstract}
 Deep learning based approaches have been utilized to model and generate graphs subjected to different distributions recently. However, they are typically unsupervised learning based and unconditioned generative models or simply conditioned on the graph-level contexts, which are not associated with rich semantic node-level contexts. Differently, in this paper, we are interested in a novel problem named Time Series Conditioned Graph Generation: given an input multivariate time series, we aim to infer a target relation graph modeling the underlying interrelationships between time series with each node corresponding to each time series. For example, we can study the interrelationships between genes in a gene regulatory network of a certain disease conditioned on their gene expression data recorded as time series. To achieve this, we propose a novel Time Series conditioned Graph Generation-Generative Adversarial Networks (TSGG-GAN) to handle challenges of rich node-level context structures conditioning and measuring similarities directly between graphs and time series. Extensive experiments on synthetic and real-word gene regulatory networks datasets demonstrate the effectiveness and generalizability of the proposed TSGG-GAN.
\end{abstract}

\section{Introduction}

Due to the important role of graphs in effectively and vividly modeling real-world collections of pairwise relational data, generative models for real-world graphs have found widespread applications, such as inferring gene regulatory networks, modeling social interactions and discovering new molecular structures. Just like the problem of inferring gene regulatory networks from expression data, many problems in graph generation can be posed as translating an input multivariate time series data as node expression values into a corresponding output graph, since graphs’ node-level expression time series data are easier to be obtained rather than the real unseen graph topologies. It would be highly desirable if we could develop generative graphs models that can directly learn from their corresponding expression data, which promotes the understanding of their underlying functional structures and discovery of meaningful structures with desired properties. \par
\begin{figure}[!t]
\centering
\includegraphics[width=3.2in]{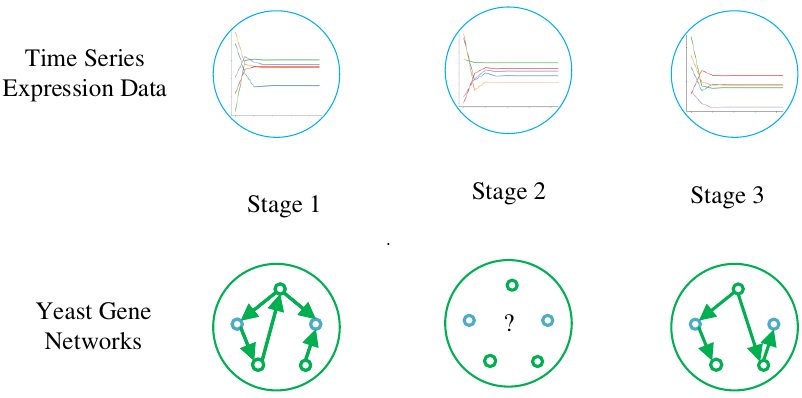}  
\caption{Toy example of inferring yeast gene regulatory networks from corresponding time series expression data. This enables us to explore the possible intrinsic patterns between node-level contexts recorded as time series and corresponding graph structures, and generate graph structures with available easy-obtained time series expression data.}
\label{fig_1}
\end{figure}

In this work, we propose and study the novel problem of time series conditioned graph generation, which aims to learn and generate graph topologies given the node-level expression data recorded as time series. Figure \ref{fig_1} demonstrates an example of inferring gene regulatory networks from corresponding expression data. Given paired graphs and time series, a desired model should learn to generate the best likely real graph candidates for the given new unseen expression time series. The problem is quite important since it can generate graph structures which are fundamental to understand underlying interrelationships between genes. \par

Traditional approaches to tackling the problem of generating graphs from corresponding time series \cite{liu2015dynamic,kenett2010dominating,sugihara2012detecting,friedman2008sparse} have two major limitations. First, they cannot generalize to new problem instances and have to solve the instances of the same type of problem again and again. Second, during the optimization process, to measure the similarities between generated graphs and unseen ground truth graphs which generate the real expression data, they have to use generated graphs to simulate expression time series data and regard the similarities between the simulated expression data and the real time series as the surrogate similarities between corresponding generated and ground truth graphs. This surrogate similarity may introduce biases because many graphs may correspond to similar or same expression time series. It would be desirable if we can directly measure the similarity between expression time series data and graphs, without the burden of transforming them into the same domain.\par

 On the other side, the community has already taken significant steps in taking advantages of recent advances in deep generative models for generative modeling graph data, such as generative adversarial networks (GAN) \cite{goodfellow2014generative} and variational autoencoders (VAE) \cite{kingma2013auto}. Based on these approaches a large number of deep learning models for generating graphs have been proposed \cite{guimaraes2017objective,zhou2019misc,li2018learning,grover2018graphite,simonovsky2018graphvae,de2018molgan,bojchevski2018netgan,kipf2016variational,yang2019conditional,you2018graphrnn}. For example, \cite{simonovsky2018graphvae} proposed a method based on VAE towards generation of small graphs, and \cite{you2018graphrnn} proposed a deep auto-regressive models for realistic graphs generation. However, these recently proposed deep models are either limited to modeling a single graph \cite{grover2018graphite,de2018molgan,kipf2016variational,bojchevski2018netgan}, or modeling sets of graphs belonging to the same semantic class \cite{jin2018junction,li2018learning,simonovsky2018graphvae,you2018graphrnn}. Only three research works study the problem of conditional graph generation, but the conditions in their setting are direct given graph properties, such as number of nodes \cite{kipf2016variational}, nodes’ categorical labels \cite{fan2019labeled} and graph-level contexts \cite{yang2019conditional}. None of the existing methods study the problem of learning to translate complex time series domain into another complex graph domain.\par

 	In this paper, to address the essential challenges of rich node-level contexts conditioning and directly measuring the similarity between time series and graphs, we explore GANs in the conditional setting and propose the novel model of TSGG-GAN for time series conditioned graph generation. Specifically, the generator in a TSGG-GAN adopts a variant of recurrent neural network called Simple Recurrent Units (SRU) \cite{lei2018sru} to extract essential information from time series, and outputs weighted matrices representing graphs. Moreover, to directly measure the similarity between time series and graphs, the discriminator fully leverages the well-developed graph convolutional neural networks (GCN) \cite{kipf2016semi} to transform the weighted graphs into node embeddings and permutation-invariant graph embeddings, with the latent representations of time series serving as node features, and then compare the graph embeddings to the latent time series’ representations to calculate their similarities. Our approach is the first deep generative method that addresses the problem of graph-structured data generation conditioned on time series expression data. \par

 To fully demonstrate the value of time series conditioned graph generation and the effectiveness of our proposed TSGG-GAN model, extensive experiments are conducted on synthetic and real-world gene regulatory networks of varying sizes and characteristics, showing TSGG-GAN achieves superior quantitative performance through careful comparisons over various graph properties.

\section{Related Work}
\paragraph{GANs for graph generation.} In recent years, there has been a surge of research in applying GANs on graph-structured data generation. For example, \cite{guimaraes2017objective} proposed ORGAN based on GANs and reinforcement learning to guide the generation of molecules encoded as text sequences. Also based on GANs and reinforcement learning, \cite{de2018molgan} introduced MolGAN, an implicit, likelihood-free generative model for small molecular graphs generation. \cite{bojchevski2018netgan} proposed NetGAN to learn the distribution of biased random walks over the input graph from which graph structure can be inferred. \cite{zhou2019misc} proposed Misc-GAN to model the underlying distribution of graph structures at different levels of granularity and transferred them into target graphs. \cite{fan2019labeled} proposed LGGAN to train deep generative models for graph-structured data with node label and can generate diverse labeled graphs. \cite{yang2019conditional} proposed CONDGEN to address the problem of conditional structure generation based on VAE and GANs. \par

\paragraph{Siamese neural network for similarity measuring.} Siamese neural networks have been widely used for similarity metric learning in various domains. For example, \cite{bromley1994signature} proposed a siamese time delay neural network for signature verification. \cite{zagoruyko2015learning} trained convolutional neural networks for learning a similarity function which can be used to compare image patches. \cite{pei2016modeling} studied siamese recurrent networks to minimize a classification loss for learning a good similarity measure between time series. \cite{bai2019simgnn} designed a siamese GCN-based model to learn a function mapping a pair of graphs into a similarity score.

\section{TSGG-GAN}
We first describe the problem definition of time series conditioned graph generation, and then present our conditional generative framework, TSGG-GAN.

\subsection{Problem Formulation}

This paper focuses on the novel problem of time series conditioned graph generation, translating an input multivariate time series to a target graph. We are provided with a set of paired data $(ts, g)=\{(ts_1, g_1),(ts_2, g_2), \dots, (ts_n, g_n)\}$, where $g_i = (V_i, E_i, A_i)$ represents a directed weighted graph described by the set of nodes $v_i$, the set of directed edges $E_i$ and the set of edge weights $A_i$. $ts_i$ denotes the set of $i$th input multivariate time series, with each time series recording the expression data of corresponding node in $g_i$. \par

In this work, we focus on learning the translation from the time series domain to the graph space, with the aim to infer the underlying graph topology representing the interrelationships between nodes from their expression data recorded as time series. Specifically, we define this new problem as time series conditioned graph generation, where we focus on learning a translation mapping $T: ts_x \rightarrow g_y$ from an input multivariate time series $ts_x \in \mathbb{TS}_X$ to a target directed weighted graph  $g_y \in \mathfrak{g}_Y$, where $\mathbb{TS}_X$ and  $\mathfrak{g}_Y$ represents the domains of input multivariate time series and target weighted graphs, respectively.

\subsection{TSGG-GAN: Time Series Conditioned Generative Models for Graphs}
\begin{figure}[!t]
\centering
\includegraphics[width=3.2in]{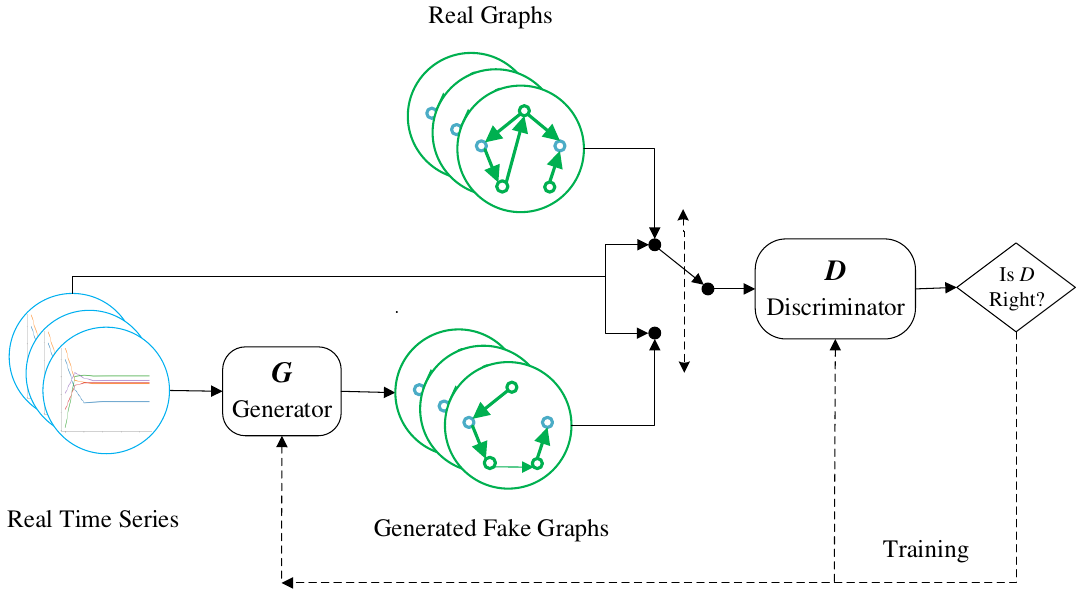}  
\caption{Overall architecture of TSGG-GAN. Generator and discriminator are specially designed for time series conditioned graph generation, with the aim to fully leverage the node-level contexts in inferring graph structures.}
\label{fig_2}
\end{figure}
To address the novel problem defined above, we propose TSGG-GAN, which leverages the joint power of SRU, GCN, and GAN for time series conditioned graph generation. Figure \ref{fig_2} demonstrates the overall architecture of proposed TSGG-GAN, consisting of two main components: a conditioned generator $G$ and a conditioned discriminator $D$. Generator $G$ and discriminator $D$ operate as two opponents: generator $G$ tries to fit $p_{true}(g|ts)$, learning a translation mapping from the input multivariate time series $ts$ to the graph distribution $p(g)$ for sampling new graphs, while discriminator $D$ learns a good similarity measurement between graphs and time series, classifying whether graph candidate came from the true graph distribution rather other from generator $G$. $G$ and $D$ are both implemented as neural networks, playing the following two-player minimax game with value function $V(G, D)$:

\begin{align}\label{eq:1}
     \mathop {\min }_G \mathop {\max }_D V(D,G) =& {\mathbb{E}_{g \in {p_{data}}(g)}}[\log D(g,ts)]+ \nonumber\\
    + &{\mathbb{E}_{z \in {p_z}(z)}}[\log (1 - D(G(z,ts),ts))]
\end{align}%

\subsection{Generator}

A defining feature of time series to graph translation problems is that they map a set of multivariate time series with various kinds of characteristics to complex graph topologies. In addition, unlike traditional image-to image translation problems whose input and output are both images, for the problems we consider, the input and output both represent totally different modalities. Therefore, we need to best utilize the information conveyed in the input time series, and learn to generate graphs with each node roughly representing a certain time series. The generator architecture is designed in specific purpose around these considerations.\par

  The TSGG-GAN’s generative model adopts SRU to fast extract essential information best likely representing the time series and uses a multi-layer perceptron (MLP) to generate the directed weighted graph. The generator $G$ takes as input the latent representation of the multivariate time series after passing the SRU model, and produces one weight matrix $A \in R^{N \times N}$, which denotes the weighted connections among nodes in a graph, with each node corresponding to each time series. Different from previous works regarding the output of $G$ as an unweighted probability matrix \cite{fan2019labeled,yang2019conditional,simonovsky2018graphvae}, we directly use it as the final weight matrix for a certain graph.

\subsection{Discriminator}
\begin{figure}[!t]
\centering
\includegraphics[width=3.2in]{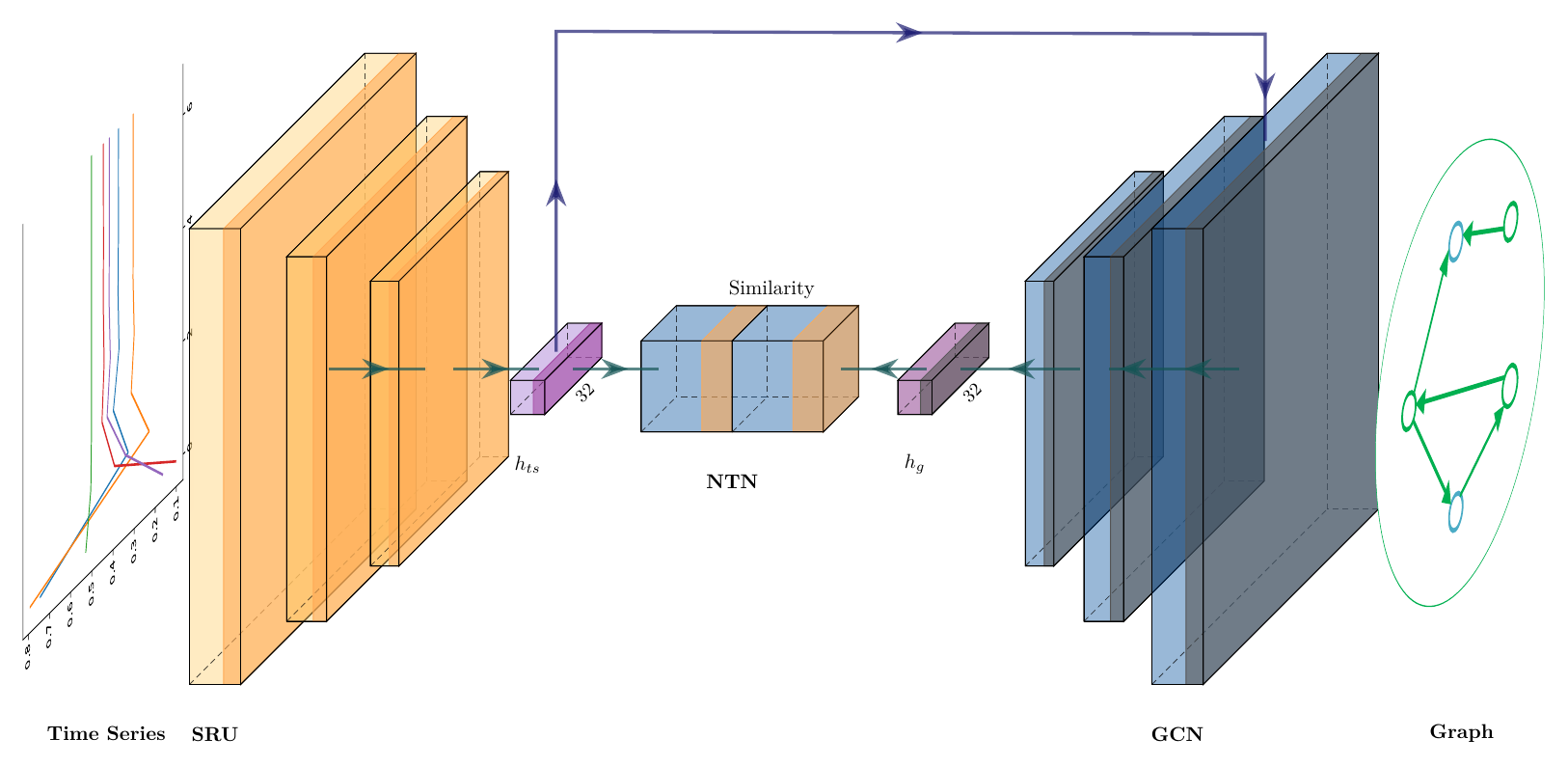}  
\caption{Overall framework of discriminator. The left part is a SRU, where we extract essential information from input time series and form a single latent representation of time series. The right part is a GCN, where we transform the input graph into a graph embedding. When generating the graph embedding using GCN, the latent representation of time series severs as node features, so as to fully leverage rich node-level contexts.}
\label{fig_3}
\end{figure}

The discriminator $D$ tries to distinguish the “fake” graphs from “real” graphs which most likely represent the input multivariate time series. $D$ receives a pair of a graph and a multivariate time series as input and outputs a single scalar measuring their similarity. \par

To address this problem, as shown in Figure \ref{fig_3}, we propose a pseudo-siamese neural network architecture that enables to measure the similarity directly between a graph candidate and a multivariate time series, without the need of first translating the graph into the time series space by simulation and then calculating the difference between these two time series. Our model $D$ adopts SRU to extract meaningful information from multivariate time series, and utilizes the GCN model to deal with weighted graphs, which is composed of a series of graph convolutional layers and an aggregation layer to extract powerful structure-aware graph representations. For a weighted graph with $N$ nodes denoted as $A \in R^{N \times N}$, and with $d_1$-dimensional nodes features denoted as $X \in R^{N \times d_1}$, a GCN model calculates the graph’s node embeddings as follows:\par
\begin{equation}\label{eq:2}
h(A,X) = \tilde{A}{\rm{ReLU(}}\tilde{A}X{W_0}){W_1}
\end{equation}
where $h(A,X)$ is a 2-layer GCN model, $W_0 \in R^{d_1 \times d_2}$ and $W_1 \in R^{d_2 \times d_3}$ are parameters in the first and second layer, respectively. $\tilde A = {D^{ - \frac{1}{2}}}A{D^{ - \frac{1}{2}}}$ is the symmetrically normalized adjacent matrix of $A$, where $D$ is its degree matrix. For the weighted graphs, we calculate their degree matrices after discarding the edges whose absolute value of weight is less than 0.05 according to \cite{stach2012learning}. \par

After two layers of node embedding propagation via GCN, similar to \cite{li2015gated}, the node embeddings are aggregated into a graph-level representation vector as:\par
\begin{equation}\label{eq:3}
  {h_g} = \sum\limits_{v \in } {\sigma (i(} {h_v}^T)) \odot \tanh (j({h_v}^T))
\end{equation}
where  $\sigma(\cdot)$ is the sigmoid function, $i$ and $j$ are linear layers and  $\odot$ means element-wise multiplication. Different from \cite{li2015gated}, we do not use tanh activation function in the last layer to restrict the graph embedding in the range of [-1, 1]. \par

Before every graph convolutional layer and the final aggregation layer, feature representations of nodes are concatenated with latent representation of the input multivariate time series, which is obtained by passing through a SRU model, leading to better capture of underlying relationships between latent space of time series and latent space of graphs. Given the latent representations of multivariate time series and the graph-level embeddings of graphs produced by the above strategy, a simple way to measure their similarity is to pass them through an MLP model and output a scalar value. However, as studied in \cite{socher2013reasoning}, such simple strategy for measuring the relationship between time series and graphs tends to be weak and insufficient. Thus we use their proposed Neural Tensor Networks (NTN) to model the similarity relationship between time series and graphs:\par
\begin{equation}\label{eq:4}
  s({h_{ts}},{h_g}) = f({h_{ts}}^T{W^{[1:K]}}{h_g} + V \begin{bmatrix}
                                                               h_{ts}  \\
                                                               h_g
                                                             \end{bmatrix}{l}
 + b)
\end{equation}
where $h_{ts}$ denotes the latent representation of multivariate time series, $h_g$ is the graph-level embedding of graphs, ${W^{[1:K]}} \in {R^{D \times D \times K}}$ and $V \in {{\rm{R}}^{K \times 2D}}$  are both weight matrices, $b$ is a bias vector, and $f(\cdot)$ is the $\tanh$ activation function. $K$ is a hyperparameter determining the number of similarity scores produced by $D$ model for each pair of latent representations of time series and graph-level embeddings of graphs.
\subsection{Optimization and Inference}

	We follow the standard training strategy from \cite{goodfellow2014generative}, alternately training one step on $D$, then two steps on $G$, and we use minibatch SGD with batch size of one. In addition, training GANs to work well is known to be painful and unstable, therefore we adopt several techniques to effectively optimize our neural networks.\par
\paragraph{Better loss function.}We adopt the least squares loss function advised by [28] to improve stability of learning process.
\begin{align}\label{eq:5}
     \mathop {\min }_D V(D) =& \frac{1}{2} {\mathbb{E}_{g \in {p_{data}}(g)}}[ (D(g,ts) - 1)^2] + \nonumber\\
      + & \frac{1}{2} {\mathbb{E}_{z \in {p_{z}}(g)}}[ (D(z,ts) - 1)^2] \nonumber\\
    \mathop {\min }_G V(G) = &{\mathbb{E}_{z \in {p_z}(z)}}[(D(G(z,ts),ts))^2]
\end{align}%

\paragraph{Better training objective for $G$.} Besides the loss function defined in Eq. \ref{eq:5} for $G$, we also utilize the feature-matching strategy to add two additional objectives for $G$.
\begin{align}\label{eq:6}
    \mathop {\min }_G V(G) =& \frac{1}{2}{E_{z \in {p_z}(z),ts}}[{(D(z,ts))^2}] + \nonumber \\
    + & \alpha \| h_{g_{real}} - h_{g_{z \in {p_z}(z)}} \|_2 + \nonumber \\
    +& \beta \| h_{t{s_{real}}} - h_{t{s_{z \in {p_z}(z)}}}\|_2+ \nonumber \\
    + &\omega \|ts_{real} - ts_{z \in {p_z}(z)}\|_2
\end{align}%
where  $h_{g_{real}}$ and $h_{g_{z \in {p_z}(z)}}$ represent the graph-level embeddings of real graphs and fake graphs generated by $G$, and  $h_{t{s_{real}}}$ and $h_{t{s_{z \in {p_z}(z)}}}$ represent the latent representations of corresponding real time series and fake time series simulated by fake graphs, respectively. \par
\paragraph{Better optimizer.} We use a recently proposed optimizer called RAdam, a new variant of Adam, to optimize models and use the default hyperparameters suggested by the authors, with the exception of learning rate being 0.0002 for generator and 0.0001 for discriminator. RAdam uses the learning rate warmup as a variance reduction technique, achieving remarkable success in stabilizing training, accelerating convergence and improving generalization.
\section{Experiments}
In this section, we conduct extensive experiments on time series conditioned graph generation and compare TSGG-GAN to state-of-the-art baselines to demonstrate its effectiveness and scalability to generate high-quality time series conditioned graphs in diverse settings.\par

\subsection{Datasets}
We perform experiments on both synthetic and real datasets, with varying sizes and characteristics. The synthetic dataset is constructed based on scale-free graphs, consisting of three subsets with different graph sizes: 10, 50 and 100. Each subset has 800 pairs of input time series and target graphs: 400 paired time series and graphs are used for training TSGG-GAN while the remaining 400 pairs are used as test dataset. For constructing the corresponding paired time series for given graphs, a simulation model is required to obtain expression time series data. Particularly, in this work, we adopt the fuzzy cognitive maps \cite{kosko1986fuzzy} to generate graphs’ corresponding time series.\par
\begin{equation}\label{eq:7}
  {T_g}^i(t + 1) = \psi (\mathop \sum \limits_{j = 1}^N {A_{ij}}{T_g}^j(t))
\end{equation}
where ${T_g}^i(t)$ is the expression value of node $i$ at the $t$th iteration of graph $g$, $A_{ij}$ stands for the relationship originating from the $j$th node and pointing to the $i$th node, and  $\psi(\cdot)$ is the sigmoid activation function.\par

\paragraph{Scale-free Graphs.} 800 graphs with $|V|$ = {10, 50, 100} are generated using the Barab\'{a}si-Albert model, with 0.4 as the probability for adding an edge between two existing nodes \cite{stach2012learning}. Since the generated graphs of BA model are unweighted, they are transformed to weighted graphs simply by setting the weights sampled uniformly from [-1, 1] to existing edges. Corresponding multivariate time series data are generated using Eq. \ref{eq:7}, with initial node values are sampled uniformly from [0, 1].\par

\paragraph{Real World Benchmarks: DREAM3.}  DREAM3 datasets come from the in \emph{silico} network challenge \cite{greenfield2010dream4,marbach2009generating,stolovitzky2007dialogue}, which aims to infer the directed unsigned gene regulation network topologies from the available in \emph{silico} gene expression datasets simulated by continuous differential equations with noise and perturbation. DREAM3 contains three groups of gene networks with 10, 50 and 100 genes, and each group in DREAM3 has 5 different networks. In the following experiments on DREAM3, in the sequence of each gene node, only the last 11 time points are used, due to the fact that previous time points are observed under perturbations. Each gene network has more than one set of time series, and we report the average result generated by each set of time series.

\subsection{Experimental Setup}
\paragraph{Baselines.} Since there is no existing work on the novel problem of time series conditioned weighted graph generation using deep learning, we carefully adapt one state-of-the-art graph-level graph generation method called CONDGEN \cite{yang2019conditional}, by setting the latent representation of time series as the graph-level context and setting the ouputs of  CONDGEN being directed weighted graphs. We also compare the proposed TSGG-GAN to traditional methods for network structures estimation from time series, i.e., PCI \cite{kenett2010dominating}, MAGA \cite{liu2015dynamic}. Unlike these traditional methods which need to solve the same optimization differing in the data again and again, our model can be trained once and generalize directly to unseen instances. \par
\paragraph{Generator architecture.} The generator architecture is fixed for all experiments. The generator includes a 2-layer bidirectional SRU model of 32 hidden units and a 2-layer MLP of [32, 64] hidden units, respectively, with leaky ReLU as activation function. Finally, the last layer of MLP is linearly projected to match the corresponding graph sizes, and we further use a tanh activation function to scale the graphs’ weights into [-1, 1]. The instance normalization strategy \cite{ulyanov2016instance} is used after each linear layer except for the output layer, due to the use of the batch size of 1.
\paragraph{Discriminator architecture.} The discriminator architecture is also fixed for all experiments. The discriminator includes 2-layer bidirectional SRU model of 32 hidden units and a 2-layer GCN encoder (see Eq. \ref{eq:2}) of [32, 32] hidden units. Then a 32-dimensional graph-level representation is computed by passing the node embedding through a 2-layer MLP of dimensions [32, 64] and with leaky ReLU as hidden layer activation function. The instance normalization strategy is also used here like in the generator. Further, an NTN model with output dimension $K$=16 is used to calculated 16 similarity scores between time series and graphs.
\begin{table}[ht]
\renewcommand{\arraystretch}{1.3}
\centering
\begin{tabular}{lllll}
\toprule
 Methods & Datasets  & HIM &  QJSD \\
\midrule
       & BA-10   & 0.41 & 0.12    \\
PCI    & BA-50  & 0.42 & 0.32    \\
       & BA-100   & 0.42 & 0.52    \\
        & DERAM3  & 0.63 & 0.28     \\
\hline
       & BA-10    & 0.39 & 0.12    \\
MAGA  & BA-50  & 0.32 & 0.21    \\
       & BA-100   & 0.33 & 0.31     \\
        & DERAM3  & 0.59 & 0.25     \\
\hline
       & BA-10   & 0.34 & 0.07    \\
CONDGEN    & BA-50   & 0.27 & \textbf{0.18}    \\
       & BA-100   & 0.27 & 0.30     \\
       & DERAM3  & 0.57 & 0.17     \\
\hline
       & BA-10  & \textbf{0.25} & \textbf{0.06}   \\
TSGG-GAN & BA-50   & \textbf{0.26} & \textbf{0.18}    \\
       & BA-100   & 0.28 & \textbf{0.26}     \\
        & DERAM3  & \textbf{0.55} & \textbf{0.14}     \\
\bottomrule
\end{tabular}
\caption{Performance evaluation over compared algorithms regarding several graph distance functions. Smaller values indicate higher similarities to the real graphs}
\label{tab:1}
\end{table}
\paragraph{Hyperparameters setting.}  In the loss function of $G$ (see Eq. \ref{eq:6}), three hyperparameters   are set to 1, 0.5 and 50, respectively. For each experiment, the TSGG-GAN is trained for total 100 epochs. And the trained model on each group of synthetic dataset is used to infer real-word datasets.

\subsection{Performance}
To measure the similarities between generated graphs and ground truth graphs, following existing works on graph generative models \cite{guo2018deep}, the generated graphs are compared with the ground truth graphs through several graph distance methods, namely, combination of Hamming and Ipsen-Mikhailov distances (HIM) \cite{jurman2015him}, and spectral entropies of the density matrices (QJSD). \par

 The set of graph statistics we use measure the similarity between the generated graphs and the ground truth from different perspectives. As shown in Table 1, our proposed TSGG-GAN outperforms all other compared algorithms and constantly ranks top except for one experiment.   The advantage of TSGG-GAN is that TSGG-GAN can capture the rich information hidden in the multivariate time series with different characteristics and produce graphs best likely mimicking the ground truth graph. Moreover, TSGG-GAN learns a direct distance metric between graphs and time series, which makes it not heavily influenced by the biases introduced by the surrogate time series loss function. Finally, once trained, TSGG-GAN can be applied to new problem instances and generate corresponding graphs without further optimization, while traditional algorithms have to optimize each problem instance repeatedly one by one.
%
%

%
%
%
%
%
%
%
%

\section{Conclusion}
We proposed TSGG-GAN, a conditional graph generative model to infer graph topologies from given time series expression data. To address the two unique challenges of rich node-level context-structure conditioning and measuring similarity directly between the domain of graphs and the domain of time series, we designed TSGG-GAN by the joint power of SRU, GCN and GAN models. Extensive experiments on both synthetic and real-world gene regulatory network dataset demonstrate the effectiveness and generalizability of TSGG-GAN compared to previous state-of-the-art models. We hope our work would inspire following-up research on addressing the remaining problems of generating various size of graphs, instead of fixed number of nodes when training TSGG-GAN, and scaling to larger graphs.
%
%

\bibliographystyle{named}
\bibliography{ijcai20-multiauthor}

\end{document}